\documentclass{article}

\usepackage{arxiv}
\usepackage[utf8]{inputenc} 
\usepackage[T1]{fontenc}    
\usepackage{hyperref}       
\usepackage{url}            
\usepackage{booktabs}       
\usepackage{amsfonts}       
\usepackage{nicefrac}       
\usepackage{microtype}      
\usepackage{lipsum} 
\usepackage{multirow}
\usepackage{graphicx}
\usepackage{caption}
\usepackage{subcaption}
\usepackage{moresize}
\usepackage{float}

\newcommand{\Description}[2][]{}
\title{Draw your Neural Networks}

\author{
  Jatin ~Sharma \\
  Microsoft Research\\
  Redmond, WA 98052 \\
  \texttt{jatin.sharma@microsoft.com} \\
  \And
  Shobha ~Lata \\
  Microsoft Corporation\\
  Redmond, WA 98052 \\
  \texttt{shobha.lata@microsoft.com} \\
}

\begin{document}
\maketitle

\begin{abstract}
Deep Neural Networks are the basic building blocks of modern Artificial Intelligence. They are increasingly replacing or augmenting existing software systems due to their ability to learn directly from the data and superior accuracy on variety of tasks. Existing Software Development Life Cycle (SDLC) methodologies fall short on representing the unique capabilities and requirements of AI Development and must be replaced with Artificial Intelligence Development Life Cycle (AIDLC) methodologies. In this paper, we discuss an alternative and more natural approach to develop neural networks that involves intuitive GUI elements such as blocks and lines to draw them instead of complex computer programming. We present Sketch framework, that uses this GUI-based approach to design and modify the neural networks and provides interoperability with traditional frameworks. The system provides popular layers and operations out-of-the-box and could import any supported pre-trained model making it a faster method to design and train complex neural networks and ultimately democratizing the AI by removing the learning curve.
\end{abstract}


\section{Introduction}
\label{sec:introduction}
Deep Neural Networks have high expressibility and learn directly from the data without any feature engineering. This makes them highly versatile and the foundational blocks for the modern Artificial Intelligence. These networks comprise of dozens and sometimes hundreds of self-learning layers stacked one after another. Each of these layers have weights, biases and several other learnable parameters that are adjusted through the back-propagation algorithm \cite{Rumelhart1986}\cite{LeCun1988} during the network’s training phase to learn useful patterns. Creation of these networks is generally complex and requires expertise in computer programming, mathematics and application-specific domain. For example, deep neural networks for computer vision generally deploy convolutional layers \cite{LeCun1998} which look at spatial patterns, however, the networks for time-series analysis or language modeling require recurrent layers (e.g. RNN \cite{Rumelhart1986}\cite{Jordan1986}\cite{Sherstinsky2020}, LSTM \cite{Hochreiter1997}\cite{Sherstinsky2020}, BERT \cite{Devlin2019}) which gather temporal patterns and long-term dependencies. A network for video processing may use both convolutional and recurrent layers. Some systems may deploy attention \cite{Vaswani2017} based layers that look out for correlations between input and desired output. Therefore, it requires a high learning curve, in-depth technical background and years of experience before making any meaningful contribution in designing novel architectures. There exist numerous deep learning frameworks such as PyTorch \cite{Paszke2019}, Tensorflow \cite{Abadi2016}, Keras \cite{Chollet2018}, Caffe \cite{Jia2014}, Torch \cite{torch}, MXNet \cite{Chen2015} and so on that provide extensive capabilities to design and train neural networks but they are programming-based and are generally not interoperable. Recently, ONNX \cite{bai2019} project has initiated a formal discussion and collaboration across the industry to encourage interoperability between these platforms. There is also a lack of change management system that could track the modifications in the existing networks and provide a version control such that it becomes possible to go back and forth between the generations of the trained neural networks.

This paper proposes Sketch framework that try to bridge this gap in the technology and provides an alternative approach for the AI development. Instead of computer programming which requires language specific syntax and structure, the system uses intuitive GUI elements such as blocks and lines to represent layers and their interconnections. Conceptually, it is similar to flow-charts where a user can just draw a neural network instead of programming it and the resulting neural network is a visual sketch instead of a long monolithic piece of code. Thus providing an intuitive and much faster approach to neural network development. The system can also import and export models from major deep learning frameworks (e.g. PyTorch, Torch, ONNX) providing a standard interface for interoperability. Most of the applications require some standard layers (e.g. convolution layer, fully-connected layer, pooling, non-linearity), loss functions (e.g. mean square error, absolute loss) and other operations (e.g. batch normalization, dropout, identity). The proposed system come up with these standard layers and operations out-of-the-box and thus provides a much faster method to design and train complex neural networks without the technical complexities.

\section{Related Work}
\label{sec:related_work}
Computer programming based neural network development has been the prominent approach in AI-Development so far. There are several deep learning frameworks which are publicly available. Some of them include PyTorch \cite{Paszke2019}, Tensorflow \cite{Abadi2016}, Keras \cite{Chollet2018}, Caffe \cite{Jia2014}, Torch \cite{torch} and MXNet \cite{Chen2015}. PyTorch and Tensorflow has been the most popular for their ease-of-use and large community support. Another leading reason for the popularity has been attributed to their support to Python as primary development language which is widely used in the research community due to its user-friendly syntax and faster learning curve. These frameworks have their custom implementations and generally lack interoperability which presents a huge roadblock moving from prototyping to production phase. In this direction, Open Neural Network Exchange (ONNX) \cite{bai2019} is the first multi-organization attempt to encourage framework-interoperability and shared optimization. Today more than 40 organizations support the ONNX initiative which highlights the necessity for interoperability in AI. 
 
However, there has not been significant development in GUI-based AI development and interoperability by design. There exist some well supported graphical interfaces like Caffe GUI Tool \cite{CaffeGuiTool}, Nvidia Digits \cite{Digits}, Sony Neural Network Console \cite{Sony}, Mathwork nnstart \cite{nnstart} and Expresso \cite{Expresso2015} but they lack capabilities of a comprehensive framework. The authors in \cite{klemm2018} developed Barista as an Open Source tool for designing and training neural networks. However, it is based on Caffe and therefore limited in the user base. Researchers at MIT developed Elegant Neural Network User Interface (ENNUI) \cite{ennui} - a browser-based neural network editor that presents a drag and drop interface to create the models and export to python code. However, this also lacks general usability through the ability to import pre-existing and pre-trained models and export to interoperable formats. Recently, another tool -- Deep Recognition \cite{DeepCognition} -- is launched as the first commercially available GUI based editor for neural network development which provides a more comprehensive set of functionality including support for autoML and multi-GPU training however it is not open source and limited to PyTorch and Keras only. We believe that all these tools are paving a path towards a comprehensive graphical interface based neural network editor that could cater the needs of the AI-Development Life Cycle.

\section{AI Development Life Cycle}
\label{sec:ai_development_life_cycle}

Software Development Life Cycle (SDLC) is a well established methodology in the field of computer science. It recommends clearly defined processes for creating high quality software. With the rise of Artificial Intelligence (AI) and its increasing applications in software development, progress towards an AI-Development Life Cycle (AIDLC) is natural and roughly follows the same underlying principles as SDLC. However, it also poses unique challenges in 1) Method of development, 2) Interoperability, 3) Explainability and Debugging, 4) Collaborative Development and 5) Version Control and Maintenance. A comprehensive AI-Development framework should provide basic functionality to support these areas. In our current work, we focus on Method of development, Interoperability and Maintenance. We develop Sketch\footnotemark -- a graphical interface based AI development framework where a user can draw neural networks instead of writing complex computer code. The system supports importing existing model architectures along with their pre-trained weights to further improve them by training or benchmarking against new datasets. The system also supports a plug and play kernel based approach where the user-drawn model is stored as an abstract graph representation and could be converted to any framework of choice including PyTorch, Torch (Lua) and ONNX. Thus, providing interoperability by design. 

\footnotetext{Please visit \url{https://github.com/jatinsha/sketch} for further details and code related to Sketch project.}

\section{System Description}
\label{sec:system_description}

\begin{figure}[]
    \centering
    \includegraphics[width=1\linewidth]{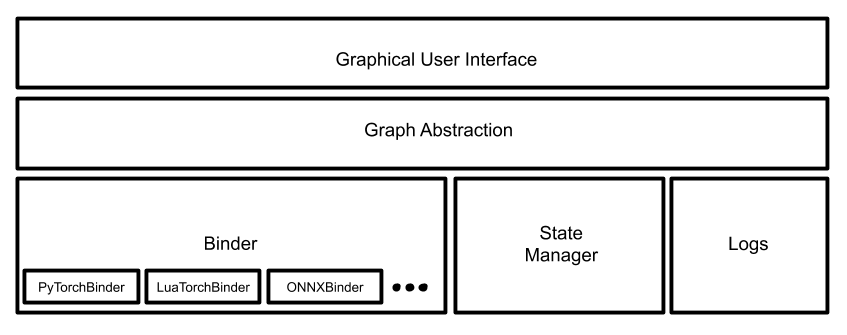}
    \captionsetup{justification=centering,margin=0cm}
    \caption{A block diagram illustrating the major components of the Sketch system}
    \label{fig:block_diagram}
    \Description[]{}
\end{figure}

The proposed system has five major components as depicted in figure \ref{fig:block_diagram}. Next, we would discuss each of these components in detail.

\subsection{Graphical User Interface}
\label{sec:graphical_user_interface}

\begin{figure}[]
    \centering
    \includegraphics[width=1\linewidth]{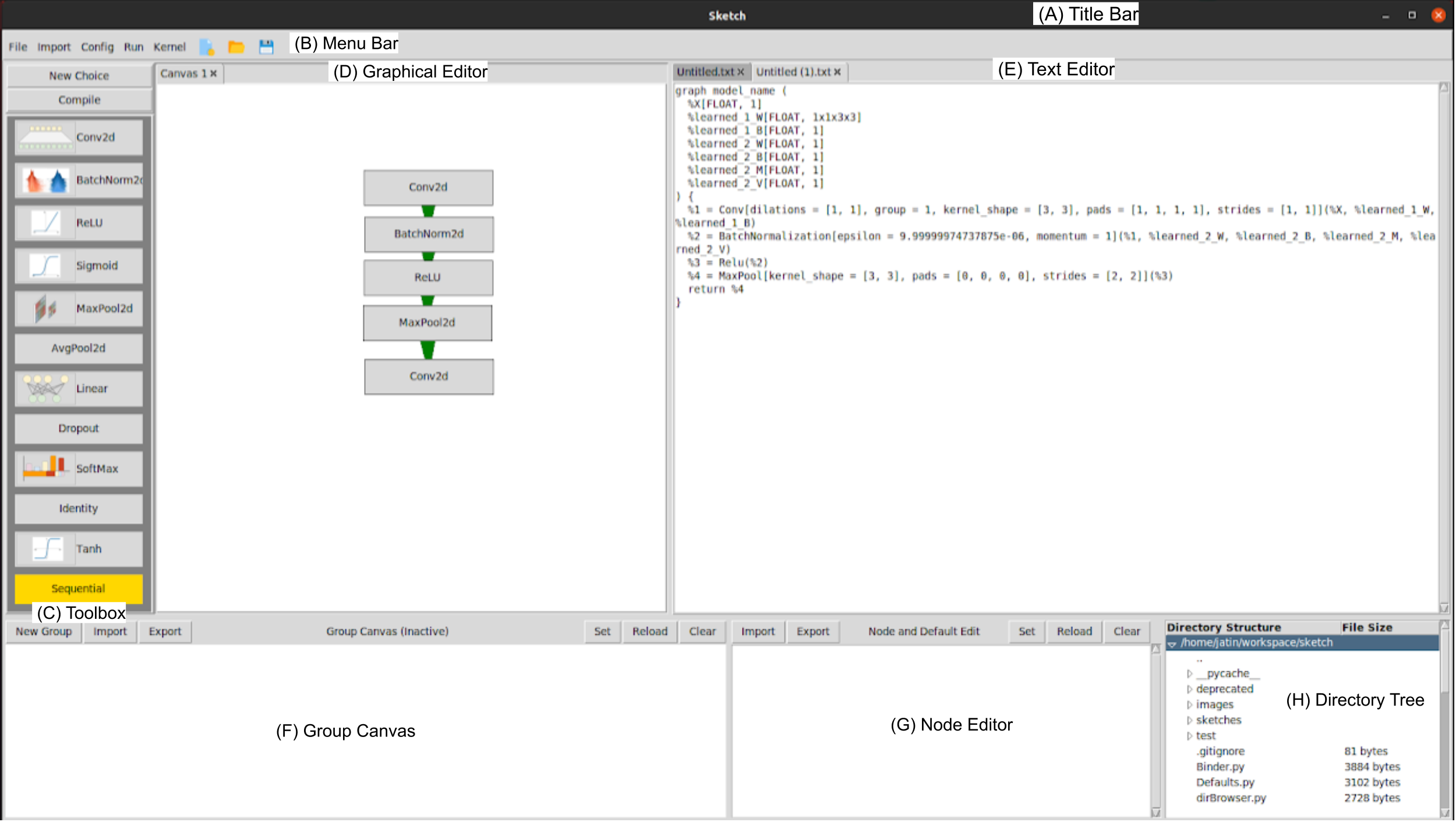}
    \captionsetup{justification=centering,margin=0cm}
    \caption{A typical example of Sketch's Graphical User Interface}
    \label{fig:Sketch_GUI}
    \Description[]{}
\end{figure}

This component provides the primary interaction between the user and the system. It encompasses all the UI elements that help in drawing the neural networks and compiling them. Figure \ref{fig:Sketch_GUI} demonstrates a screenshot of the system's typical graphical user interface. At the very top, there is Title Bar (A) which has minimize, maximize and close buttons along with the name of the tool. Beneath it lies the Menu Bar (B) which contains various menu options to import/export models, open/save files,  choose target kernel and update/reset configuration settings. Toolbox (C) is located at the left side and it contains popular layers. Layers from the toolbox can be used through a ‘Drag \& Drop’ method to a Canvas on the Graphical Editor (D). These individual network layers which look like rectangular blocks could be connected by clicking on their edges and dragging the link to another layer and thus forming arrow-like inter-connections.

When a network is compiled with the help of a kernel, an output model file is generated which contains both model architecture and its parameters. A representation text output is populated in the Text Editor (E). The Sequential layer is used to create a group of layers or custom computation blocks that could further be edited through Group Canvas (F). The parameters (e.g. channels, kernel size, stride, padding) of any individual layer could be modified by clicking at a layer on the canvas and editing the corresponding values in the Node Editor (G). Located on the bottom right corner, Directory Tree (H) displays an interactive list of current working directory and can be used to open any file. 

\subsection{Graph Abstraction}
\label{sec:graph_abstraction}
   
\begin{figure}[H]
    \centering
    \includegraphics[width=1\linewidth]{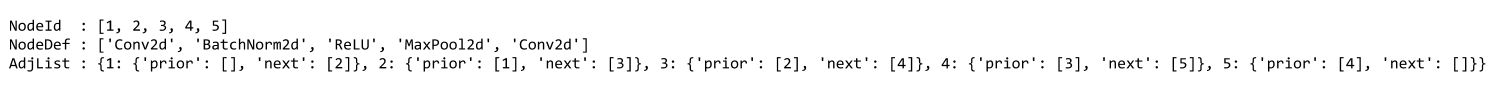}
    \captionsetup{justification=centering,margin=0cm}
    \caption{An example of adjacency list representation used in Sketch's Graph Representation}
    \label{fig:adjacency_list}
    \Description[]{}
\end{figure}

A user draws a neural network on a canvas in the Graphical Editor using various layers from the Toolbox and linking them together through inter-connections. To create an inter-connection, the user can click anywhere on the border of a layer and drag it to another layer. As the user draws the network, an abstract Graph representation of the same it additively created by the Graph Abstraction component. To accomplish this an Adjacency List representation is maintained where the layers and their inter-connections are used as graph nodes and edges. Each node contains its nodeId and lists for prior and next connections. An extensive set of processing happens in the background for various events such as key bindings for shortcuts, selection and grouping of multiple elements, their positioning and deletion -- making it intuitive for the user to work with the system. Figure \ref{fig:adjacency_list} illustrates an example adjacency list for the network shown in figure \ref{fig:Sketch_GUI}.

\subsection{Binder}
\label{sec:binder}

The binder is the most crucial component of the Sketch system. It acts similar to a compiler and converts the high-level abstract graph representation of the deep neural network into a framework-specific model architecture. The choice of target framework is defined through a kernel which could be changed by the user at any time. For an instance, if we draw a sequential neural network containing Convolution \textrightarrow BatchNormalization \textrightarrow ReLU \textrightarrow MaxPool \textrightarrow Convolution layers and pick PyTorch as the kernel then compiling the canvas would generate a PyTorch model as *.pth file along with a textual PyTorch representation in textual editor. If we switch the kernel to ONNX and re-compile the same canvas then it would output an *.onnx model and ONNX representation of the underlying neural network. 

The Binder interface exposes two methods -- exportModel and importModel. The exportModel method provides basic capability to convert the Graph Abstraction into framework-specific representation. On the other hand, importModel enables loading any pre-existing model from the supported frameworks and converting them into the system’s Graph Abstraction and a sketch on the canvas. This provides a tremendous capability to this tool to not only create new model architectures but also to use and modify huge number of pre-trained models available in the research community. This Binder interface is implemented by various kernels which provide the actual low-level conversion logic and weight transfer between the frameworks. The Sketch system currently supports PyTorch, Torch (Lua) and ONNX through implementing corresponding kernels. Similar kernels could be implemented for TensorFlow, Caffe or any other framework pretty easily as a plug and play method.

\subsection{State Manager}
\label{sec:state_manager}

\begin{figure}[]
    \centering
    \includegraphics[width=1\linewidth]{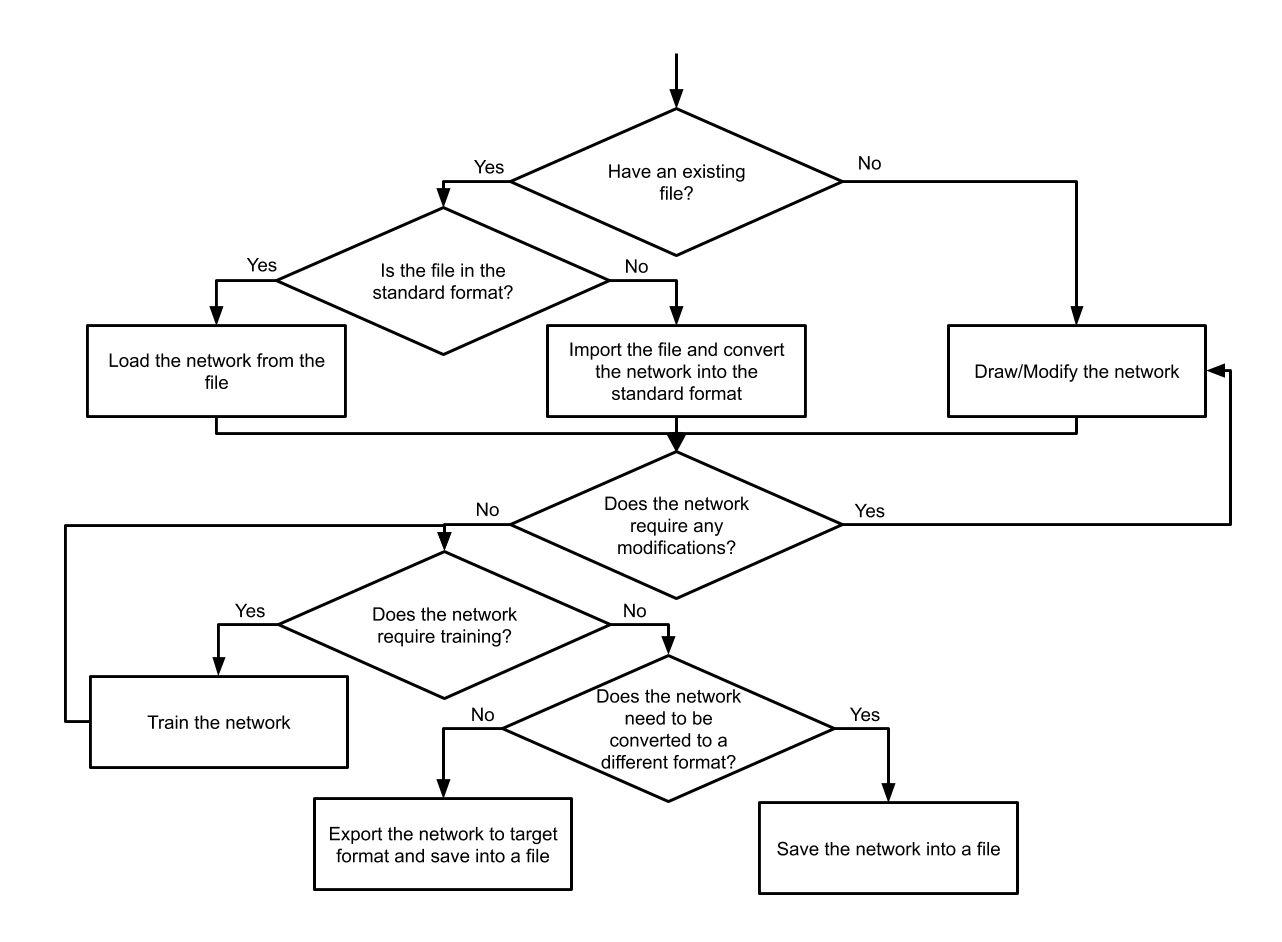}
    \captionsetup{justification=centering,margin=0cm}
    \caption{A flowchart diagram illustrating the key steps in Sketch-based AI development}
    \label{fig:flow_diagram}
    \Description[]{}
\end{figure}

The State Manager is responsible for tracking the system state. At all time, the crucial state information such as system version, current working directory path and list of open tabs, canvas state and so on is tracked. When the editor is closed this state information is saved to the disk and reloaded whenever the edited is reopened. Using this mechanism, Sketch editor try to restore the open files and canvases which were being worked on in the previous session. Every modification on the canvas is recorded by a checkpointing mechanism under the state manager and is used to provide the undo and redo functionalities. Figure \ref{fig:flow_diagram} illustrates they key steps in Sketch-based AI development.

\subsection{Logs}
\label{sec:logs}
   
To provide a debugging and maintenance mechanism crucial programmatical events (e.g. loading/saving files, changing kernel, compiling canvas) are logged. During an exception, the appropriate stack trace is logged as well. This provides a sequential trace of various actions and system state which could be used to debug and enhance the system.

\section{Conclusions}
\label{sec:conclusions}

In the present work, we have discussed the need for research towards AI-Development Life Cycle as existing Software Development Life Cycle Strategies may not completely represent the unique challenges faced by the AI development. From computer coding based method to framework-specific implementations and from trace based debugging to version control, there exist several areas that need to be reworked in accordance with the unique characteristics of the AI. We developed Sketch framework which -- (1) provides an alternative and natural approach to develop neural networks, and (2) brings interoperability by design through providing import and export capabilities for major frameworks in a plug \& play fashion. This removes dependency upon the extensive knowledge of underlying deep learning framework and computer programming language syntax and provides a lego-like functionality to build over existing operators. We believe such an approach could democratize the AI development. In future, we intend to work towards implementing In-built Training, Version Control, Collaborative Development and Visual Debugging towards a comprehensive and one-stop framework for AI Development.

\bibliographystyle{unsrt}  
\bibliography{references}

\end{document}